\documentclass[11pt,a4paper]{article}
\usepackage[hyperref]{acl2020}
\usepackage{times}
\usepackage{latexsym}
\usepackage{booktabs}
\usepackage{hyperref}
\usepackage{colortbl}
\usepackage{tablefootnote}

\newcommand{\etc}{\textit{etc.}}

\newcommand{\projname}{IndicNLP }
\newcommand{\besthl}[1]{\textbf{#1}}
\newcommand{\bulletc}{\noindent$\bullet\;$}
\newcommand{\ftlinkshow}[1]{\footnote{#1}}
\newcommand{\tblftlinkshow}[1]{\tablefootnote{#1}}
\newcommand{\ftlink}[1]{}
\newcommand{\hh}[1]{\multicolumn{1}{c}{\textbf{#1}}} 

\usepackage{microtype}

\aclfinalcopy % Uncomment this line for the final submission
\title{AI4Bharat-\projname Corpus: Monolingual Corpora and Word Embeddings for Indic Languages}

\author{Anoop Kunchukuttan$^1$, Divyanshu Kakwani$^2$, Satish Golla$^3$, Gokul N.C.$^4$, \\
\textbf{Avik Bhattacharyya$^5$, Mitesh M. Khapra$^6$, Pratyush Kumar$^7$}\\
Microsoft India$^1$, IIT Madras$^{2,6,7}$, AI4Bharat$^{3,4,5}$\\
\texttt{ankunchu@microsoft.com}$^1$, \texttt{gokulnc@ai4bharat.org}$^{4}$,\\ \texttt{\{gsatishkumaryadav,avikbhattacharyya.2k\}@gmail.com}$^{3,5}$,\\
\texttt{\{divk,miteshk,pratyush\}@cse.iitm.ac.in}$^{2, 6,7}$
}

\date{}

\begin{document}
\maketitle
\begin{abstract}
We present the \projname corpus, a large-scale, general-domain corpus containing 2.7 billion words for 10 Indian languages from two language families. We  share pre-trained word embeddings trained on these corpora. We create news article category classification datasets for 9 languages to evaluate the embeddings. We show that the \projname embeddings significantly outperform publicly available pre-trained embedding on multiple evaluation tasks. We hope that the availability of the corpus will accelerate Indic NLP research. The resources are available at https://github.com/ai4bharat-indicnlp/indicnlp\_corpus.
\end{abstract}

\section{Introduction}

Distributional representations are the corner stone of modern NLP, which have led to significant advances in many NLP tasks like text classification, NER, sentiment analysis, MT, QA, NLI, \etc $\;$  Particularly, word embeddings \cite{mikolov2013distributed}, contextualized word embeddings \cite{peters2018deep},  and language models \cite{devlin2018bert} can model syntactic/semantic relations between words and reduce feature engineering. These pre-trained models are useful for initialization and/or transfer learning for NLP tasks. They are  useful for learning multilingual embeddings, enabling cross-lingual transfer. 
Pre-trained models are typically learned using unsupervised approaches from large, diverse monolingual corpora. 
The quality of embeddings is impacted by the size of the monolingual corpora \cite{mikolov2013efficient,bojanowski2017enriching}, a resource  not widely available publicly for many major languages.

Indic languages, widely spoken by more than a billion speakers, lack large, publicly available monolingual corpora. They include 8 out of top 20 most spoken languages and $\sim$30 languages with more than a million speakers. 
There is also a growing population of users consuming Indian language content (print, digital, government and businesses). %\todo{Google-KPMG study}. %To enable digital consumption and rich user experience, NLP tools are needed for Indian languages. 

Indic languages are very diverse, spanning 4 major language families. The Indo-Aryan and Dravidian languages are spoken by  96\% of the population in India. The other families are diverse, but the speaker population is relatively small. Almost all languages have SOV word order and are morphologically rich.
%(Indo-Aryan  and Dravidian languages exhibit fusional and agglutinative morphology respectively). 
The language families have also interacted over a long period of time leading to significant convergence in linguistic features; hence, the Indian subcontinent is referred to as a \textit{linguistic area} \cite{emeneau1956india}. So, Indic languages are of great interest and importance for NLP research.

In this work, we address the creation of large, general-domain monolingual corpora for multiple Indian languages. Using the new monolingual corpora, we also create other essential resources for Indian language NLP. We evaluate these resources on various benchmarks, in the process creating some new evaluation benchmarks. This work contributes the following Indic language resources: 

\bulletc A large monolingual corpora (\projname corpus) for 10 languages from two language families (Indo-Aryan branch and Dravidian). Each language has at least 100 million words (except Oriya). 

\bulletc Pre-trained word embeddings for 10 Indic lan-guages trained using FastText.

\bulletc News article category classification datase for 9 languages.

\bulletc Unsupervised morphanalyzers for 10 languages.

We show that \projname embeddings outperform publicly available embeddings on various tasks: word similarity, word analogy, sentiment analysis, text classification, bilingual lexicon induction. Further, we show the utility of the monolingual corpora for training morphanalzyers.

\section{Related Work} 

\noindent\textbf{Text Corpora.} Few organized sources of monolingual corpora exist for most Indian languages. The EMILLE/CIIL corpus \cite{mcenery2000emille} was an early effort to build a corpora for South Asian languages, spanning 14 languages with a total of 92 million words. \textit{Wikipedia} for Indian languages is small (the largest one, Hindi, has just 40 million words). The Leipzig corpus \cite{goldhahn2012leipzig} contains small collections of upto 1 million sentences for news and web crawls (average 300K sentences). In addition, there are some language specific corpora for Hindi and Urdu \cite{bojar2014hindencorp,jawaid2014tagged}.

The \textit{CommonCrawl} project crawls webpages in many languages by sampling various websites. Our analysis of a processed crawl for the years 2013-2016 \cite{buck-etal-2014-n} for Indian languages  revealed that most Indian languages, with the exception of Hindi, Tamil and Malayalam, have few good sentences ($\geq$10 words) - in the order of around 50 million words. The Indic corpora size has not possibly changed siginificantly in recent years based on the recent CommonCrawl statistics.

\noindent\textbf{Word Embeddings.} Word embeddings have been trained for many Indian languages using limited  corpora. The Polyglot \cite{polyglot:2013:ACL-CoNLL} and FastText projects provide embeddings trained on Wikipedia. FastText also provides embeddings trained on Wikipedia + CommonCrawl corpus.

\section{\projname Corpus Creation} 

\begin{table}[t]
\setlength{\tabcolsep}{3pt} % Default value: 6pt
\centering
\begin{tabular}{lrrr}
\toprule
\hh{Lang} & \hh{\#Sentences} & \hh{\#Tokens}    & \hh{\#Types}   \\
\midrule
pa       & 6,525,312 & 179,434,326 & 594,546 \\
hi & 62,961,411& 1,199,882,012& 5,322,594 \\
bn       & 7,209,184   & 100,126,473 & 1,590,761 \\
or       & 3,593,096 & 51,592,831 & 735,746 \\
gu       & 7,889,815   & 129,780,649 & 2,439,608 \\
mr       & 9,934,816   & 142,415,964 & 2,676,939 \\
\midrule
kn      & 14,780,315   & 174,973,508 & 3,064,805 \\
te       & 15,152,277  & 190,283,400 & 4,189,798 \\
ml       &  11,665,803 & 167,411,472 & 8,830,430 \\
ta & 20,966,284 & 362,880,008 & 9,452,638 \\
\midrule
Total & 160,678,313     & 2,698,780,643 & 38,897,865   \\
\bottomrule
\end{tabular}
\caption{\projname corpus statistics}
\label{tab:mono_stats}
\end{table}

We describe the creation of the \textit{IndicNLP} corpus.  

\noindent\textbf{Data sources.} Our goal is collection of corpora that reflects contemporary use of Indic languages and covers a wide range of topics. Hence, we focus primarily on the news domain and Wikipedia. We source our data from popular Indian languages news websites, identifying many news sources from \textit{W3Newspapers}\ftlinkshow{https://www.w3newspapers.com}. We augmented our crawls with some data from other sources: Leipzig corpus \cite{goldhahn2012leipzig} (Tamil and Bengali), WMT NewsCrawl (for Hindi), WMT CommonCrawl \cite{buck-etal-2014-n}  (Tamil, Malayalam), HindEnCorp (Hindi) \cite{bojar2014hindencorp}.

\noindent\textbf{News Article Crawling.} We use \textit{Scrapy}\ftlinkshow{https://scrapy.org/}, a web-crawling Python framework, for crawling news websites. If the site has a good sitemap, we rely on it to increase the efficiency of the crawler. For other sources, we crawl all the links recursively. 

\noindent\textbf{Article Extraction.} For many news websites, we used \textit{BoilerPipe}\ftlinkshow{https://github.com/kohlschutter/boilerpipe}, a tool to automatically extract the main article content for structured pages without any site-specific learning and customizations \cite{Kohlschtter2010BoilerplateDU}. This approach works well for most of the Indian language news websites. 
In some cases, we wrote custom extractors for each website using \textit{BeautifulSoup}\ftlinkshow{https://www.crummy.com/software/BeautifulSoup}, a Python library for parsing HTML/XML documents. After content extraction, we applied some filters on content length, script \etc, to select good quality articles. We used the \textit{wikiextractor}\ftlinkshow{https://github.com/attardi/wikiextractor} tool for text extraction from Wikipedia. 
 
\noindent\textbf{Text Processing.} First, we canonicalize the representation of Indic language text  in order to handle multiple Unicode representations of certain characters and typing inconsistencies. Next, we sentence split the article and tokenize the sentences.  These steps take into account Indic punctuations and sentence delimiters. Heuristics avoid creating sentences for initials (P. G. Wodehouse) and common Indian titles (Shri., equivalent to Mr. in English) which are followed by a period. We use the \textit{Indic NLP Library}\ftlinkshow{https://github.com/anoopkunchukuttan/indic\_nlp\_library} \citep{kunchukuttan2020indicnlp} for processing. 

The final corpus collection is created after de-duplicating and shuffling sentences. We used the Murmurhash algorithm (\textit{mmh3} python library with a 128-bit unsigned hash) for de-duplication. 

\noindent\textbf{Dataset Statistics.} Table \ref{tab:mono_stats} shows statistics of the monolingual datasets for each language. Hindi and Tamil are the largest collections, while Oriya has the smallest collection. All other languages have a collection between 100-200 million words. Bengali, a widely spoken language, has only around 100 million words: we would like to increase that collection. The Hindi corpus is a compilation of existing sources. CommonCrawl is a significant contributor to the Tamil corpus (55\%) and Malayalam (35\%). Most of the data for other languages originate from our crawls.

\section{\projname Word Embeddings}

We train pre-trained word embeddings using the \projname corpus, and evaluate their quality on: (a) word similarity, (b) word analogy, (c) text classification, (d) bilingual lexicon induction tasks. We compare the \projname embeddings with two pre-trained embeddings released by the \textit{FastText} project trained on  Wikipedia (\textit{FT-W}) \cite{bojanowski2017enriching} and Wiki+CommonCrawl (\textit{FT-WC}) \cite{grave2018learning} respectively. This section describes the training of word embeddings, and evaluation settings and results for each task.

\subsection{Training Details}

We train 300-dimensional word embeddings for each language on the \projname corpora using \textit{FastText} \cite{bojanowski2017enriching}.
 Since Indian languages are morphologically rich, we chose \textit{FastText},  which is capable of integrating subword information by using character n-gram embeddings during training. 

 We train skipgram models for 10 epochs with a window size of 5, minimum token count of 5 and 10 negative examples sampled for each instance. We chose these hyper-parameters based on suggestions by \citet{grave2018learning}. Otherwise,  default settings were used.   
 Based on previously published results, we expect FastText to be better than word-level algorithms like \textit{word2vec} \cite{mikolov2013distributed} and \textit{GloVe} \cite{pennington2014glove} for morphologically rich languages. We leave comparison with word-level algorithms for future work.

\subsection{Word Similarity \& Analogy Evaluation}

\begin{table}[t]
\centering
\begin{tabular}{lrrr}
\toprule 
\hh{Lang} & \hh{FT-W} & \hh{FT-WC} & \hh{INLP} \\
\midrule
\multicolumn{4}{l}{\textbf{Word Similarity} (\textit{Pearson Correlation})} \\
pa       &  \besthl{0.467} & 0.384 & 0.428\\
hi       & 0.575 & 0.551         & \besthl{0.626} \\
gu       & 0.507 & 0.521         & \besthl{0.614} \\
mr       & 0.497 & \besthl{0.544}         & 0.495 \\
te       & 0.559 & 0.543         & \besthl{0.560} \\
ta       & \besthl{0.439} & 0.438         & 0.392 \\
\midrule
Average & 0.507 & 0.497 & \besthl{0.519} \\
\midrule 
\multicolumn{4}{l}{\textbf{Word Analogy} (\textit{\% accuracy})} \\
hi       & 19.76 &    32.93      & \besthl{33.48} \\
\bottomrule
\end{tabular}
\caption{Word Similarity and Analogy Results}
\label{tab:wordsim_results}
\end{table}

We perform an intrinsic evaluation of the word embeddings using the IIIT-Hyderabad word similarity dataset \cite{akhtar-etal-2017-word} which contains similarity databases for 7 Indian languages. The database contains similarity judgments for around 100-200 word-pairs per language. Table \ref{tab:wordsim_results} shows the evaluation results.  We also evaluated the Hindi word embeddings on the Facebook Hindi word analogy dataset \cite{grave2018learning}. \projname embeddings outperform the baseline embeddings on an average. 

\subsection{Text Classification Evaluation}

\begin{table}[t]
\setlength{\tabcolsep}{3pt} % Default value: 6pt
\centering
\begin{tabular}{llrr}
\toprule
\hh{Lang} & \hh{Classes} & \multicolumn{2}{c}{\textbf{\# Articles}}\\ 
 &  & \hh{Train} & \hh{Test} \\
\midrule
pa       & BIZ, ENT. POL, SPT  & 2,496 &  312\\
bn       & ENT, SPT  & 11,200 & 1,400 \\
or       & BIZ, CRM, ENT, SPT & 24,000 &  3,000\\
gu       & BIZ, ENT, SPT & 1,632 & 204 \\
mr       & ENT, STY, SPT  & 3,815 & 478\\
\midrule
kn      &  ENT, STY, SPT & 24,000 & 3,000\\
te       & ENT, BIZ, SPT  & 19,200 &  2,400\\
ml       & BIZ, ENT, SPT, TECH & 4,800 & 600\\
ta       & ENT, POL, SPT  & 9,360 &  1,170 \\
\bottomrule
\end{tabular}
\caption{\projname News category dataset statistics. The following are the categories: entertainment: ENT, sports: SPT, business: BIZ, lifestyle; STY, techology: TECH, politics: POL, crime: CRM}
\label{tab:indicnlp_classification_stats}
\end{table}

We evaluated the embeddings on different text classification tasks:  (a) news article topic, (b) news headlines topic, (c) sentiment classification. We experimented on publicly available datasets and a new dataset (\projname News Category dataset). 

\noindent\textbf{Publicly available datasets.}  We used the following datasets: (a) IIT-Patna Sentiment Analysis dataset \cite{akhtar16hybrid}, (b) ACTSA Sentiment Analysis corpus \cite{mukku-mamidi-2017-actsa}, (c) BBC News Articles classification dataset,  (d) iNLTK Headlines dataset,  (e) Soham Bengali News classification dataset. Details of the datasets can be found in the Appendix \ref{apx:public_classification_datasets}. Our train and test splits derived from the above mentioned corpora are made available on the \projname corpus website. 

\noindent\textbf{\projname News Category Dataset.} We use the \projname corpora to create classification datasets comprising news articles and their categories for 9 languages. The categories are determined from URL components. We chose generic categories like entertainment and sports which are likely to be consistent across websites. The datasets are balanced across classes. See Table \ref{tab:indicnlp_classification_stats} for details. 

\noindent\textbf{Classifier training.} We use a $k$-NN ($k=4$) classifier since it is a non-parameteric - the classification performance directly reflects the how well the embedding space captures text semantics \cite{meng2019spherical}. The input text embedding is the mean of all word embeddings (bag-of-words assumption). 
%A test instance ia labelled with the class that is the most common among its $k$ nearest neighbors in the training set.

\noindent\textbf{Results.} On nearly all datasets \& languages, \projname embeddings outperform  baseline embeddings (See Tables \ref{tab:classification_results1} and \ref{tab:classification_results2}). 

\begin{table}[t]
\centering
\begin{tabular}{lrrr}
\toprule 
\hh{Lang} & \hh{FT-W} & \hh{FT-WC} & \hh{INLP} \\
\midrule
pa       &  94.23 & 94.87         & \besthl{96.79} \\
bn       & 97.00 & 97.07         & \besthl{97.86} \\
or       & 94.00 & 95.93         & \besthl{98.07} \\
gu       & 97.05  & 97.54         & \besthl{99.02} \\
mr       &  96.44 & 97.07          & \besthl{99.37} \\
\midrule
kn       &  96.13 &  96.50       & \besthl{97.20} \\
te       & 98.46  &  98.17         & \besthl{98.79} \\
ml       &  90.00 &  89.33        & \besthl{92.50} \\
ta       & 95.98  &    95.81      & \besthl{97.01} \\
\midrule
Average  & 95.47   &  95.81         & \besthl{97.40} \\
\bottomrule
\end{tabular}
\caption{Accuracy on \projname News category testset}
\label{tab:classification_results1}
\end{table}

\begin{table}[t]
\setlength{\tabcolsep}{3pt} % Default value: 6pt
\centering
\begin{tabular}{llrrr}
\toprule 
\hh{Lang} & \hh{Dataset} & \hh{FT-W} & \hh{FT-WC} & \hh{INLP} \\
\midrule
hi & BBC Articles  & 72.29 & 67.44 & \besthl{74.25} \\
 & IITP+ Movie   & 41.61 & 44.52 & \besthl{45.81} \\
 & IITP Product & 58.32 & 57.17 & \besthl{63.48} \\
 \midrule
bn &  Soham Articles  & 62.79 & 64.78 & \besthl{72.50}\\
\midrule
gu & 	& 81.94 & 84.07 & \besthl{90.90}\\
ml & iNLTK 	& 86.35 & 83.65 & \besthl{93.49}\\
mr & Headlines	& 83.06 & 81.65 & \besthl{89.92}\\
ta & 	& 90.88 & 89.09 & \besthl{93.57}\\
\midrule
te & ACTSA & 46.03 & 42.51 & \besthl{48.61}\\
\midrule
 & Average  & 69.25 &  68.32 & \besthl{74.73}\\
\bottomrule
\end{tabular}
\caption{Text classification accuracy on public datasets}
\label{tab:classification_results2}
\end{table}

\subsection{Bilingual Lexicon Induction}

We use \projname embeddings for creating mutlilingual embeddings, where monolingual word embeddings from different languages are mapped into the same vecor space. Cross-lingual learning using multilingual embeddings is useful for Indic languages which are related and training data for NLP tasks is skewed across languages.  We train bilingual word embeddings from English to Indian languages and vice versa using GeoMM \cite{jawanpuria2018learning}, a state-of-the-art supervised method for learning bilingual embeddings. We evaluate the bilingual embeddings on the BLI task, using bilingual dictionaries from the MUSE project and \textit{en-te} dictionary created in-house. We search among the 200k most frequent target language words with the CSLS distance metric during inference \cite{conneau18a}. The quality of multilingual embeddings depends on the quality of monolingual embeddings. \projname bilingual embeddings significantly outperform the baseline bilingual embeddings (except Bengali). 

\begin{table}[t]
\centering
\setlength{\tabcolsep}{3pt}
\begin{tabular}{llllrrr}
\toprule
 & \multicolumn{3}{c}{\textbf{en to Indic}} & \multicolumn{3}{c}{\textbf{Indic to en}}\\
\cmidrule(lr){2-4} 
\cmidrule(lr){5-7} 
 &\hh{\small{FT-W}} & \hh{\small{FT-WC}} & \hh{\small{INLP}} & \hh{\small{FT-W}} & \hh{\small{FT-WC}} & \hh{\small{INLP}} \\
\midrule
bn     & 22.60    & \besthl{33.92}    & 33.73  & 31.22   & \besthl{42.10}    & 41.90     \\   
hi     & 40.93   & 44.35    & \besthl{48.69}  & 49.56   & 57.16    & \besthl{58.93}    \\   
te     & 21.10    & 23.01    & \besthl{29.33}    & 25.36   & 32.84    & \besthl{36.54}    \\ 
ta     & 19.27   & 30.25    & \besthl{34.43}    & 26.66   & 40.20     & \besthl{42.39}    \\ 
\midrule
Ave.  & 25.98 & 32.88 & \besthl{36.55} &	33.20 &	43.08 & \besthl{44.94}
 \\
\bottomrule
\end{tabular}
\caption{Accuracy@1 for bilingual lexicon induction}
\label{tab:bli_results}
\end{table}

\section{Unsupervised Morphology Analyzers}

Indian languages are morphologically rich. The large vocabulary poses data sparsity problems for NLP applications. Morphological analysis provides a means to factor the words into its constituent morphemes. However, morphanalyzers are either not available for many Indic languages or have limited coverage and/or low quality. Significant linguistic expertise and effort is need to build morphanalyzers for all major Indic languages. On the other hand, unsupervised morphanalyzers can be easily built using monolingual corpora.  

\noindent\textbf{Training.} We trained unsupervised morphanalyzers using Morfessor 2.0. \citep{virpioja2013morfessor}. We used only the word types (with minimum frequency=5) without considering their frequencies for training. This  configuration is recommended when no annotated data is available for tuning. 
\noindent\textbf{SMT at morph-level.} We consider SMT between Indic languages as a usecase for our morphanalzyers. We compare word-level models and two morph-level models (trained on \projname corpus and \citet{kunchukuttan2016orthographic}'s model) to verify if our morphanalyzers can address data sparsity issues. We trained Phrase-based SMT models on the ILCI parallel corpus \citep{jha2012ilci} (containing about 50k sentences per Indian language pair). We use the same data and training configuration as  \citet{kunchukuttan2016orthographic}.

\noindent\textbf{Results.}   We see that our model outperforms the word-level model significantly, while it outperforms the \citet{kunchukuttan2016orthographic}'s morphanalyzer in most cases (results in Table \ref{tab:smt_indic}). Note that the their morphanalyzer is trained on data containing the parallel corpus itself, so it may be more tuned for the task. Thus, we see that the \projname corpora can be useful for building morphological analyzers which can benefit downstream tasks.

\begin{table}[t]
\centering
\begin{tabular}{lrrr}
\toprule
\hh{Lang Pair} & \hh{word}  & \hh{morph}  & \hh{morph}  \\
 &   & \hh{(K\&B, 2016)} &  \hh{(INLP)} \\
\midrule
bn-hi      & 31.23 & 32.17      & \besthl{33.80}       \\
pa-hi & 68.96 & 71.29 & \besthl{71.77} \\
hi-ml      & 8.49  & 9.23       & \besthl{9.41}        \\
ml-hi      & 15.23 & 17.08      & \besthl{17.62}       \\
ml-ta      & 6.52  & \besthl{7.61}       & 7.12        \\
te-ml      & 6.62  & \besthl{7.86}       & 7.68        \\
\midrule
Average    & 22.84 & 24.21 & \besthl{24.57}        \\
\bottomrule
\end{tabular}
    \caption{SMT between Indian languages (BLEU scores)}
    \label{tab:smt_indic}
\end{table}

\section{Summary and Future Work}
We present the \projname corpus, a large-scale, general-domain corpus of 2.7 billion words across 10 Indian languages, along with word embeddings, morphanalyzers and text classification benchmarks. We show that resources derived from this corpus outperform other pre-trained embeddings and corpora on many NLP tasks. The corpus, embeddings and other resources will be  publicly available for research.

We are working on expanding the collection to at least 1 billion words for major Indian languages. We further plan to build: (a) richer pre-trained representations (BERT, ELMo), (b) multilingual pre-trained representations, (c) benchmark datasets for representative NLP tasks.
While these tasks are work-in-progress, we hope the availability of this corpus will accelerate NLP research for Indian languages by enabling the community to build further resources and solutions for various NLP tasks and opening up interesting NLP questions.

\bibliography{acl2020}
\bibliographystyle{acl_natbib}

\appendix

\section{Publicly Available Text Classification Datasets}
\label{apx:public_classification_datasets}
 We used the following publicly available datasets for our text classification experiments:

 (a) IIT-Patna Movie and Product review dataset  \citep{akhtar16hybrid}, (b) ACTSA Sentiment Analysis corpus \citep{mukku-mamidi-2017-actsa}, (c) IIT-Bombay Sentiment Analysis Dataset \citep{balamurali-2012-cross}, (d) BBC News Articles classification dataset,  (e) iNLTK Headlines dataset,  (f) Soham Bengali News classification corpus. The essential details of the datasets are described in Table \ref{tab:statistics_public_datasets}.

\begin{table} 
\setlength{\tabcolsep}{3pt} % Default value: 6pt
\centering
\begin{tabular}{llcrrl}
\toprule 
\hh{Lang} & \hh{Dataset} & \hh{N} & \multicolumn{2}{c}{\textbf{\# Examples}} \\
        &      &     & \hh{Train} & \hh{Test} \\
\midrule
hi & BBC Articles\tblftlinkshow{https://github.com/NirantK/hindi2vec/releases/tag/bbc-hindi-v0.1}  & 6 & 3,467 & 866  \\
 & IITP+ Movie Reviews   & 3 & 2,480 & 310 \\ 
& IITP Product Reviews\tblftlinkshow{http://www.iitp.ac.in/~ai-nlp-ml/resources.html} & 3 & 4,182 & 523 \\
 \midrule
bn &  Soham  Articles\tblftlinkshow{https://www.kaggle.com/csoham/classification-bengali-news-articles-indicnlp}  &  6 & 11,284 & 1411 \\
\midrule
gu & 	& 3 & 5,269  & 659 & \\
ml & iNLTK 	& 3 & 5,036 & 630 \\
mr & Headlines\tblftlinkshow{https://github.com/goru001/inltk}	& 3  & 9,672   & 1,210 &  \\
ta & 	& 3 & 5,346 & 669 \\
\midrule
te & ACTSA corpus\tblftlinkshow{https://github.com/NirantK/bharatNLP/releases} & 3 & 4,328 & 541 \\
\bottomrule

\end{tabular}
\caption{Statistics of publicly available datasets (N is the number of classes)}
\label{tab:statistics_public_datasets}
\end{table}

\paragraph{Some notes on the above mentioned public datasets}
\begin{itemize}
    \item The IITP+ Movie Reviews sentiment analysis dataset is created by merging IIT-Patna dataset with the smaller IIT-Bombay and iNLTK datasets.
    \item The IIT-Patna Movie and Product review datasets have 4 classes namely  postive, negative, neutral and conflict. We ignored the conflict class.
    \item In the Telugu-ACTSA corpus, we evaluated only on the news line dataset (named as telugu\_sentiment\_fasttext.txt) and ignored all the other domain datasets as they have very few data-points.
\end{itemize}

\end{document}